\documentclass[letterpaper, 10 pt, conference]{ieeeconf}
\IEEEoverridecommandlockouts     %
\newif\ifdraftmode
\draftmodefalse   %

\usepackage[T1]{fontenc}
\usepackage{cite}
\usepackage{amsmath,amssymb,amsfonts}
\usepackage{algorithmic}
\usepackage{graphicx}
\usepackage{textcomp}
\usepackage{xcolor}
\usepackage{colortbl}
\usepackage{booktabs}
\usepackage{multirow}
\usepackage{subcaption}
\usepackage{siunitx}

\ifdraftmode
  \usepackage[textsize=footnotesize,color=yellow!40]{todonotes}
  \usepackage[colorlinks=true,allcolors=blue,breaklinks=true]{hyperref}
\else
  \usepackage[disable]{todonotes}
  \usepackage[hidelinks,breaklinks=true]{hyperref} %
\fi
\usepackage[capitalise,noabbrev]{cleveref}
\crefname{figure}{Fig.}{Figs.}
\Crefname{figure}{Fig.}{Figs.}

\usepackage{xspace}
\newcommand{\method}{\mbox{\textsc{GeoWM}}\xspace}

\newcommand{\E}{\mathbb{E}}

\newcommand{\mat}[1]{\mathbf{#1}}

\newcommand{\K}{K}                      %
\newcommand{\Hset}{\mathcal{H}}          %
\newcommand{\hz}{h}                     %
\newcommand{\dep}{\mat{D}}              %
\newcommand{\anc}{\mat{A}}              %
\newcommand{\pose}{\mat{T}}             %
\newcommand{\cond}{\mat{c}}             %
\newcommand{\net}{F_\theta}             %
\newcommand{\noise}{\boldsymbol{\epsilon}}
\newcommand{\warp}{\mathcal{W}}         %
\newcommand{\hs}{\sigma}                %
\newcommand{\dmin}{d_{\min}}            %
\newcommand{\ft}{\tau}                  %

\newcommand{\TODO}[1]{}

\definecolor{best}{RGB}{198,239,206}
\definecolor{second}{RGB}{255,242,204}
\definecolor{oursbg}{gray}{0.90}
\definecolor{lead}{RGB}{0,66,140}
\definecolor{blockbg}{RGB}{228,236,246}
\definecolor{blocktxt}{RGB}{70,70,70}
\definecolor{dsbg}{gray}{0.94}

\newif\ifanonymous
\anonymousfalse   %

\title{\LARGE \bf
GeoWM: Efficient Direct World Modeling in Explicit Geometry
}

\ifanonymous
  \author{} %
\else
  \author{Mehrdad Noori, Guile Wu, Sam Hosseini, and Dongfeng Bai\\
          Huawei Noah's Ark Lab, Canada}
\fi

\begin{document}

\maketitle
\thispagestyle{empty}
\pagestyle{empty}          %

\begin{abstract}
Modeling 3D scene geometry and its evolution over time is essential for autonomous driving and robotics.
A common paradigm is to use world models to predict future images or latent representations of the environment and subsequently recover geometry from these predictions.
However, this paradigm does not explicitly model geometric structure and typically relies on recursive rollouts to reach longer prediction horizons, leading to error accumulation and increasing computational cost.
To address these limitations, we present \method, a geometry world model that directly forecasts future scene geometry at specified future horizons without recursive rollout.
The key idea is to leverage a geometry foundation model to transform observed RGB frames into a geometric history, which conditions a flow-matching transformer to predict the scene geometry at a specified future horizon.
We further show that a lightweight camera-motion predictor can accurately estimate the future viewpoint, and that projecting the observed geometry into the predicted viewpoint provides an effective geometric prior for future geometry forecasting.
Extensive experiments on four datasets spanning urban driving, aerial flight, and dynamic manipulation demonstrate that \method outperforms the evaluated world models in forecasting depth, camera pose, and 3D scene geometry, while substantially reducing inference time at longer horizons.

\end{abstract}

\section{Introduction}
\label{sec:intro}

Understanding the 3D structure of the environment and how it evolves over time is crucial for tasks such as autonomous driving and robotic navigation.
Recent advances in geometry foundation models~\cite{dust3r,vggt,pi3,mapanything} have shown impressive capabilities in modeling 3D scene geometry and camera parameters from images.
However, these models are inherently observational as they focus on reconstructing the present scene within the given images rather than predicting its future evolution.
As a result, they lack the ability to forecast how the scene will evolve as the camera and objects move, which is crucial for planning and control.

\begin{figure}[t]
  \centering
  \includegraphics[width=\columnwidth,trim=8.05 5.20 8.04 5.33,clip]{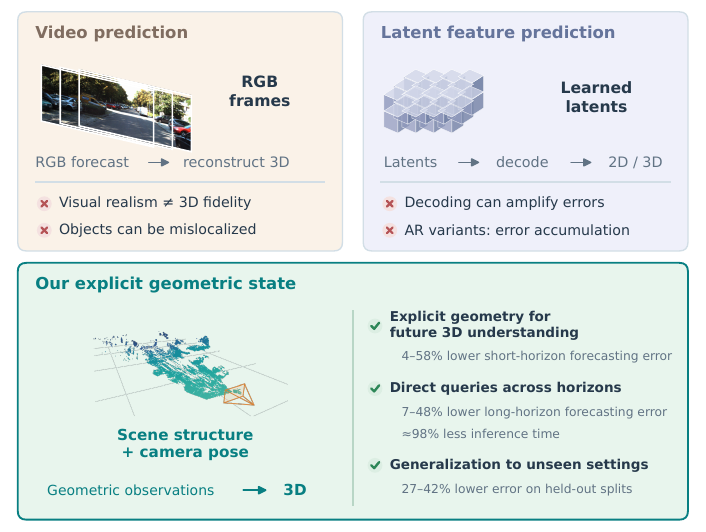}
  \caption{\textbf{Pixels, features, or explicit geometry?}
  Video models forecast future frames and then reconstruct geometry; their photometric objective couples only loosely to 3D structure, and dynamic objects can be mislocalised~\cite{vggtworld}.
  Feature-space models forecast latents and decode them, where decoding can amplify errors~\cite{vggtworld} and autoregressive rollout accumulates them~\cite{scheduledsampling,diffusionforcing,selfforcing}.
  \method forecasts explicit geometry directly and queries each horizon independently.
  }
  \label{fig:teaser}
\end{figure}

On the other hand, world models have shown promise in predicting future states of the environment as images or latent representations, from which scene geometry can subsequently be inferred.
DINO-Foresight~\cite{dinoforesight} forecasts the semantic features of a frozen vision foundation model and attaches task heads to read depth from them.
Video generators such as Cosmos~\cite{cosmos3} synthesise future RGB frames, from which geometry is reconstructed afterwards.
VGGT-World~\cite{vggtworld} moves closer to geometry, forecasting the feature states of a frozen geometry foundation model and decoding them with its pretrained heads.
Although these approaches demonstrate the potential of world models for geometry forecasting, they struggle with representing complex 3D geometry accurately (\cref{fig:teaser}).
Moreover, existing methods often employ an autoregressive rollout strategy, feeding each prediction back as the input to the next, which makes maintaining long-term consistency over extended horizons challenging as errors accumulate over time and computational cost increases.

In this work, we propose \method, a geometry world model that combines explicit geometry modeling with direct future prediction at specified future horizons.
Specifically, we employ a geometry foundation model to extract geometric information from observed frames, forming a geometric history that serves as the condition for future geometry prediction.
We then use a simple yet effective flow-matching diffusion transformer to predict the future geometry based on the observed geometric history, the predicted camera motion, and the requested future horizon.
Each requested future horizon is predicted directly, without the need for recursive rollouts, which mitigates error accumulation and reduces computational cost.
Furthermore, we employ a lightweight camera-motion predictor to estimate the future viewpoint, which is then used to project the observed geometry into the predicted viewpoint, providing an effective geometric prior for future geometry forecasting.
In this way, \method integrates explicit geometry modeling, direct future prediction, and camera-motion estimation into a cohesive framework for accurate and efficient 3D scene forecasting.

We evaluate \method on four datasets, ranging from urban driving and aerial flight to robotic manipulation, at different horizons.
We compare against three recent state-of-the-art world models: the video-based Cosmos-3~\cite{cosmos3}, the latent-geometry VGGT-World~\cite{vggtworld}, and the feature-based DINO-Foresight~\cite{dinoforesight}.
Across three families of metrics, future depth, future camera pose, and future 3D point clouds, \method outperforms other methods across horizons.
It lowers depth error by 4--58\% at short horizons and 7--48\% at long horizons against the strongest comparable baseline, with about 98\% less inference time.
Its predicted 3D point clouds are also the most accurate on all four datasets.
Together with controlled ablations of each design choice, these results show that forecasting directly in explicit geometry is both more accurate and more efficient than forecasting in pixel or feature space, and they offer a practical route to world models for the many robotic tasks that depend on 3D understanding.

In summary, our contributions are:
\begin{itemize}
  \item We introduce a geometry world model that forecasts future 3D scene geometry in an explicit geometry space extracted from observed RGB frames, providing a structured and interpretable representation for future prediction.
  \item We develop a horizon-conditioned diffusion transformer that predicts the requested future geometry directly without recursive rollouts, mitigating error accumulation, so the cost of a forecast does not grow with the horizon.
  \item Our method is efficient and outperforms previous state-of-the-art world models in future geometry forecasting at different horizons on driving, aerial, and robotic manipulation datasets, demonstrating the potential of explicit geometry as the state of a world model for tasks that depend on 3D understanding.
\end{itemize}

\section{Related Work}
\label{sec:related}

\noindent\textbf{Geometry Foundation Models.}
Feed-forward reconstruction offers an efficient alternative to optimisation-based pipelines for obtaining 3D structure from images.
DUSt3R~\cite{dust3r} showed that a single network can regress point maps, and hence depth and relative camera pose, from unposed image pairs, and VGGT~\cite{vggt} extended this to many views with a transformer that predicts camera parameters, depth maps, and point maps in one forward pass.
Subsequent models improve the formulation, for example by removing the dependence on a reference view~\cite{pi3} or by accepting metric priors and calibration as optional inputs~\cite{mapanything}.
These models describe the frames they are given with high accuracy, and we use one of them, frozen, to turn observed RGB frames into geometry.
They are not designed to predict how that geometry will change; providing this is the goal of the present work.

\noindent\textbf{Video World Models.}
Large video generators such as Cosmos~\cite{cosmos1,cosmos3} and Wan~\cite{wan}, and driving-specific models such as GAIA-1~\cite{gaia1} and Vista~\cite{vista}, are used as world models by rolling out future frames, from which geometry can then be reconstructed.
Geometry-aware variants generate depth or point maps jointly with the frames~\cite{aether,gen3r,wvd} or align the generator's features with those of a geometry model~\cite{geometryforcing}.
In both cases the forecaster remains a video model whose capacity is spent on appearance, and geometry is obtained either after generation or as one of its outputs.
Many video models generate bounded clips and extend them through autoregressive feedback, and the mismatch this creates between training on real frames and generating from predicted ones is treated as a training problem: scheduled sampling~\cite{scheduledsampling} and Self Forcing~\cite{selfforcing} expose the model to its own generated context during training, and Diffusion Forcing~\cite{diffusionforcing} trains on noise-corrupted context so that imperfect history is seen at training time.
These remedies reduce drift but keep the sequential mechanism and its cost, which grows with the horizon.

\noindent\textbf{Latent World Models.}
A second family forecasts in learned visual feature spaces.
DINO-Foresight~\cite{dinoforesight} and DINO-World~\cite{dinoworld} predict the future features of a frozen DINOv2 encoder and read out depth or segmentation with separately trained task heads, and V-JEPA~2~\cite{vjepa2} trains encoder and predictor jointly for action anticipation and planning.
DINO-World conditions on absolute timestamps, so a future time can be queried directly as well as reached autoregressively.
Most recently, VGGT-World~\cite{vggtworld} moved the latent space to a geometry foundation model: it predicts the tokens of a frozen VGGT and decodes them with the backbone's own pretrained heads, so the forecast is tied directly to camera parameters, depth maps, and point maps.
This is cheaper and closer to geometry than pixel space, but the geometry is still decoded from the forecast rather than being the forecast.
Latent models also roll out in the same way as video models: DINO-Foresight predicts features one step at a time, and VGGT-World predicts chunks of tokens, feeds them back, and relies on a flow-forcing curriculum to limit the resulting drift.
Compared with VGGT-World, our method forecasts explicit geometric quantities and predicts each requested horizon directly, without recursive feedback, using predicted camera motion to condition the geometry forecast.

\section{Method}
\label{sec:method}

\begin{figure*}[t]
  \centering
  \includegraphics[width=0.92\textwidth]{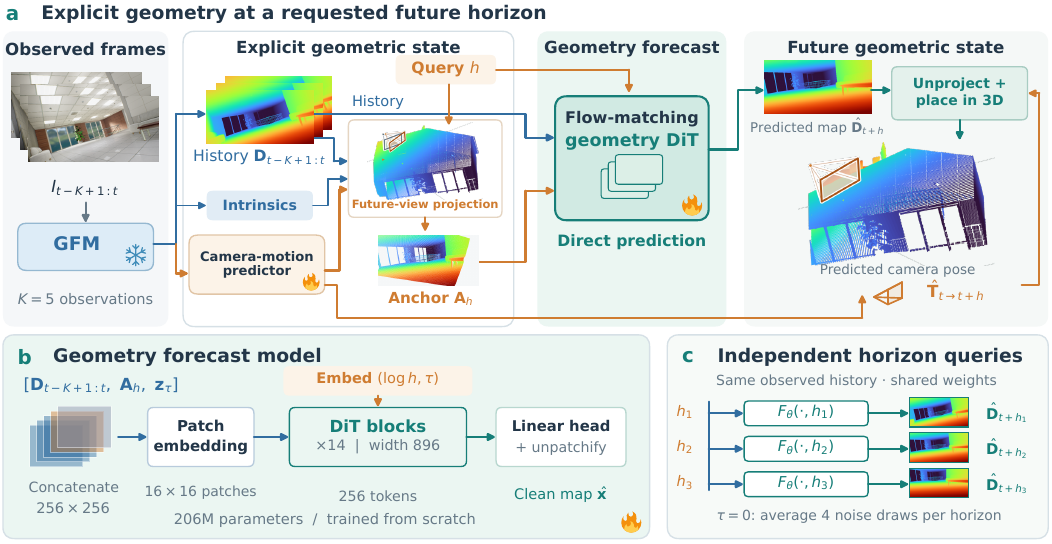}
  \caption{Overview of \method: (a) forecasting in explicit geometry, (b) the flow-matching transformer, and (c) independent horizon queries.}
  \label{fig:architecture}
\end{figure*}

\method forecasts future scene geometry directly in the output space of a geometry foundation model (\cref{fig:architecture}).
Given the $\K$ most recent RGB frames, a frozen reconstruction backbone provides a depth map and a camera pose for each frame.
A lightweight predictor extrapolates the observed camera motion to any requested horizon $\hz$, and the last observed depth map is warped into that future viewpoint to serve as geometric conditioning.
A flow-matching diffusion transformer, conditioned on the depth history, this warped depth map (the \emph{anchor}), and $\hz$, then predicts the depth map at time $t+\hz$ directly, without recursive feedback.
The horizon is thus an input to the model, and the anchor gives that input a physical meaning: it is the observation seen from where the camera is predicted to be at $t+\hz$.

\subsection{Problem Formulation}
\label{sec:problem}
Let $I_{t-\K+1:t}$ denote the $\K$ most recent frames.
A frozen reconstruction backbone maps them to depth maps $\dep_{t-\K+1:t}$, camera intrinsics, and camera poses, from which we take the $\K-1$ relative motions between consecutive frames.
Given a positive horizon $\hz$ from a dataset-defined set $\Hset$, the task is to predict the depth map $\dep_{t+\hz}$ as seen from the camera at time $t+\hz$, and the pose $\pose_{t\to t+\hz}$ of that camera relative to the last observed one, the transform that takes points from the camera frame at $t$ into the one at $t+\hz$.
We predict depth rather than point maps because, for a given camera, the pixel already fixes the viewing ray, so depth is the only free quantity per pixel; the point map follows from the predicted depth and pose.
Every horizon is predicted from the same observed history, so no prediction at a shorter horizon is used as input for a longer one.

\subsection{Camera-Motion Prediction and Geometric Conditioning}
\label{sec:anchor}
\noindent\textbf{Camera-motion predictor.}
The $\K-1$ observed relative motions are encoded as rotation vectors and normalised translation features.
A lightweight predictor maps this history directly to the relative camera poses $\hat{\pose}_{t\to t+\hz}$ of every future step in $\Hset$ at once.
We use ridge regression for rotation and a small regressor for translation; the predictor has at most a few tens of thousands of parameters.
Its accuracy is evaluated in \cref{sec:experiments} against the camera forecasts of the baselines.

\noindent\textbf{Geometric conditioning.}
Given $\hat{\pose}_{t\to t+\hz}$ and the backbone's intrinsics, we forward-project $\dep_t$ into the predicted camera with nearest-depth visibility handling, an operation we write $\warp$.
The result,
\begin{equation}
  \anc_{\hz} = \warp\big(\dep_t,\, \hat{\pose}_{t\to t+\hz}\big),
  \label{eq:anchor}
\end{equation}
is the last observation seen from the predicted future viewpoint.
The anchor supplies an estimate of the observed static structure in the future camera frame.
We use it as conditioning only: the network predicts the complete future depth map, and it learns to account for errors in the observed depth and in the predicted motion, for scene dynamics, and for regions that become visible only in the future view.

\subsection{Horizon-Conditioned Flow-Matching Transformer}
\label{sec:transformer}
\noindent\textbf{Architecture.}
The network is a plain diffusion transformer~\cite{dit} on depth maps.
Its conditioning stack is $\cond_{\hz} = (\dep_{t-\K+1:t}, \anc_{\hz})$, the observed depth maps and the anchor.
These $\K+1$ channels and the noisy target are stacked at $256\times256$, split into $16\times16$ patches, and linearly projected to 256 tokens with a learned positional embedding.
Fourteen pre-norm transformer blocks of width 896 follow, and a linear head with unpatchification gives the output.
Spatial conditioning is by channel concatenation only; the horizon and the flow time enter through the normalisation layers, as described next.
The flow time $\ft$ and $\log \hz$ are each sinusoidally embedded, concatenated, mapped by a small MLP, and injected into every block by adaptive layer normalisation with zero-initialised modulation (adaLN-zero~\cite{dit}), so that the horizon modulates every layer.
The transformer has 206M parameters and is trained from scratch; it operates directly on depth maps, with no learned autoencoder or tokenizer and no pretrained weight other than the frozen backbone.

\noindent\textbf{Objective.}
We adopt a flow-matching formulation~\cite{flowmatching,rectifiedflow} with clean-target prediction and a weighted L1 loss in velocity space.
Let $\mat{x} = \dep_{t+\hz}/\hs_{\hz}$ be the normalised future-depth target, where $\hs_{\hz}$ is a fixed per-horizon normalisation constant computed on the training set.
Let $\noise \sim \mathcal{N}(\mat{0}, \mat{I})$ be Gaussian noise and $\ft \in [0,1]$ the flow time.
The interpolant is
\begin{equation}
  \mat{z}_\ft = \ft\,\mat{x} + (1-\ft)\,\noise,
  \label{eq:interp}
\end{equation}
so that $\ft=0$ is pure noise and $\ft=1$ is the target.
The network predicts the clean target, $\hat{\mat{x}} = \net(\mat{z}_\ft, \ft, \hz, \cond_{\hz})$~\cite{jit,pointdit}, and the loss is applied to the velocity that $\hat{\mat{x}}$ implies.
Writing $x_p$ for the value of a map at pixel $p$, and taking the expectation over training windows, horizons, $\ft$, and $\noise$,
\begin{equation}
  \mathcal{L} = \E\Big[\frac{1}{\sum_p w_p}\sum_p w_p \,\Big|\frac{\hat{x}_p - z_{\ft,p}}{\max(1-\ft,\, 0.05)} - (x_p - \epsilon_p)\Big|\Big].
  \label{eq:loss}
\end{equation}
The per-pixel weight $w_p \propto m_p/\max(\dep_{t+\hz,p}, \dmin)$ is inverse depth over the valid-pixel mask $m$, bounded by a depth floor $\dmin$.
It aligns the L1 objective with relative depth error, so that nearby structure is not swamped by the far field.
A related term is used as an auxiliary loss in~\cite{pointdit}; here it is the primary weight.
Flow times follow a logit-normal distribution, $\ft = \operatorname{sigmoid}(u)$ with $u \sim \mathcal{N}(-0.8,\, 0.8^2)$~\cite{sd3}.
Following~\cite{pointdit}, we set a fixed fraction of training samples to $\ft=0$ to explicitly train the pure-noise state used at inference.

\subsection{Training and Inference}
\label{sec:training}
\noindent\textbf{Training.}
Each training example is an observed window together with a horizon $\hz$ drawn uniformly from $\Hset$; the target is the ground-truth depth at $t+\hz$.
The same network is therefore supervised at every horizon it will later be asked for, always from observed inputs.
The transformer is trained from scratch on a single GPU in a few hours.

\noindent\textbf{Inference.}
A single forward pass at $\ft=0$ from pure noise returns a clean estimate, as in the one-step inference of~\cite{pointdit}; we average $N=4$ such passes with independent noise and invert the normalisation,
\begin{equation}
  \hat{\dep}_{t+\hz} = \frac{\hs_{\hz}}{N}\sum_{n=1}^{N} \net(\noise_n, 0, \hz, \cond_{\hz}).
  \label{eq:readout}
\end{equation}
For the weighted L1 objective in \cref{eq:loss}, evaluation at $\ft=0$ reduces the training loss to the inverse-depth-weighted L1 error between $\hat{\mat{x}}$ and $\mat{x}$, so this read-out targets the relative-error criterion used by the metrics.
Averaging over noise draws reduces the remaining dependence on $\noise$, and no ODE integration is needed.
\cref{sec:experiments} compares this read-out with integrated samples.
Each requested horizon needs a fixed number of network evaluations, independent of how far ahead it lies, and queries at different horizons can be batched; unlike a rollout, no intermediate future is computed or fed back.
The predicted camera pose $\hat{\pose}_{t\to t+\hz}$ is reported as is.
For a static camera, as on DOMINO, the anchor coincides with $\dep_t$ and is omitted, so the network is conditioned on the $\K$ depth maps alone.

\section{Experiments}
\label{sec:experiments}

\subsection{Datasets and Evaluation Protocol}
\label{sec:exp:setup}

\noindent\textbf{Datasets.}
To show that \method is robust across domains, we evaluate on four challenging datasets that span driving, aerial flight, and robotic manipulation.
In every dataset, a window consists of $\K=5$ observed frames, and the horizon set $\Hset$ is fixed per dataset.
KITTI~\cite{kitti} provides 10\,Hz driving sequences with LiDAR ground truth.
We train on the 138 training drives of the KITTI depth benchmark and test on its 13 validation drives (3,257 windows), with stride 2 and 20 horizons from 0.2\,s to 4.1\,s.
Cityscapes~\cite{cityscapes} provides 17\,Hz stereo sequences.
We train on the training split and evaluate on 500 validation windows, with stride 3 and five horizons from 0.18\,s to 0.88\,s.
TartanAir~\cite{tartanair} is a challenging aerial dataset of simulated flights through 18 photorealistic environments with fast 6-DoF camera motion.
We hold out 37 of its 369 trajectories (7,314 windows) and predict 20 horizons of 1 to 20 frames.
DOMINO~\cite{domino} is one of the most challenging manipulation datasets.
Episodes are observed from a static head camera, and objects move independently of the robot.
It covers 35 tasks and five robot embodiments.
Since the public release has RGB only, we re-render the stored trajectories in SAPIEN~\cite{sapien} to obtain ground-truth depth.
To measure generalisation beyond the training distribution, we train on 23 tasks with one embodiment held out and test on four disjoint sets: seen tasks with seen robots (in-domain), novel tasks, a novel robot, and both novel (37,360 windows, 157,325 window--horizon pairs), with seven horizons from 0.24\,s to 7.7\,s.

\noindent\textbf{Metrics.}
We evaluate three aspects of the forecast: the depth map, the camera motion, and the resulting 3D geometry.
Depth is evaluated with the absolute relative error (AbsRel) at the target frame.
Following VGGT-World~\cite{vggtworld}, predictions are median-scaled per image on KITTI, Cityscapes, and TartanAir.
Since manipulation requires depth at its true scale, on DOMINO we evaluate the raw output of our method.
On KITTI, we average within each drive and then over the 13 drives.
Camera motion is evaluated with the absolute trajectory error (ATE), relative translation error (RTE), and relative rotation error (RRE) after Umeyama alignment, as in VGGT-World.
For 3D evaluation we follow the protocol of Gen3R~\cite{gen3r}: predicted and ground-truth clouds are aligned with a similarity transform, voxel-downsampled at 5\,mm and subsampled to 20k points each, and we report the mean of accuracy and completeness as the Chamfer distance (CD).

\begin{table*}[t]
\caption{\textbf{Future geometry forecasting on four datasets.}
AbsRel~($\times 100$) and $\delta_1$~(\%) score the forecast depth, CD~(m) the forecast 3D geometry after similarity alignment, and ATE~(m), RTE~(m), RRE~(deg) the forecast camera trajectory.
\emph{Next} is the first predicted horizon; \emph{Short}, \emph{Mid}, and \emph{Long} average the horizons given in each block, and \emph{All} averages every horizon.
Depth is median-scaled per image except on DOMINO, which is evaluated at true scale.
\textbf{Best} and \underline{second best} among copy-last, VGGT-World~\cite{vggtworld}, Cosmos-3~\cite{cosmos3}, and \method; copy-last repeats the last observed geometry.
}
\label{tab:main}
\centering
\fontsize{8.5}{10}\selectfont
\setlength{\tabcolsep}{1.0pt}
\renewcommand{\arraystretch}{1.04}
\newcolumntype{C}{>{\hspace{1.2pt}}c<{\hspace{1.2pt}}}
\begin{tabular}{@{}l ccC ccC ccC ccC ccC  ccc@{}}
\toprule
\multirow{2}{*}{Method} & \multicolumn{3}{c}{Next} & \multicolumn{3}{c}{Short} & \multicolumn{3}{c}{Mid} & \multicolumn{3}{c}{Long} & \multicolumn{3}{c}{All} & \multicolumn{3}{c}{Pose (trajectory-level)} \\
\cmidrule(lr){2-4}\cmidrule(lr){5-7}\cmidrule(lr){8-10}\cmidrule(lr){11-13}\cmidrule(lr){14-16}\cmidrule(lr){17-19}
& AbsRel\,$\downarrow$ & $\delta_1$\,$\uparrow$ & CD\,$\downarrow$ & AbsRel\,$\downarrow$ & $\delta_1$\,$\uparrow$ & CD\,$\downarrow$ & AbsRel\,$\downarrow$ & $\delta_1$\,$\uparrow$ & CD\,$\downarrow$ & AbsRel\,$\downarrow$ & $\delta_1$\,$\uparrow$ & CD\,$\downarrow$ & AbsRel\,$\downarrow$ & $\delta_1$\,$\uparrow$ & CD\,$\downarrow$  & ATE\,$\downarrow$ & RTE\,$\downarrow$ & RRE\,$\downarrow$ \\
\midrule
\rowcolor{dsbg}\multicolumn{19}{@{}l@{}}{\textbf{\textsc{KITTI}}\hspace{6pt}\textcolor{blocktxt}{\scriptsize next 0.21\,s, short 0.41--1.03\,s, mid 1.24--2.07\,s, long 2.27--4.14\,s}}\\[0.5pt]
Copy-last & 9.1 & 90.1 & 0.754 & 13.9 & 82.6 & 0.900 & 18.9 & 75.7 & 1.152 & 22.5 & 71.8 & 1.406 & 19.2 & 75.8 & 1.209 & 6.82 & 1.33 & 0.83 \\
DINO-Foresight & 15.0 & 81.0 & -- & 16.1 & 79.1 & -- & 19.2 & 72.1 & -- & 22.6 & 64.1 & -- & 20.0 & 69.9 & -- & -- & -- & -- \\
VGGT-World & 7.8 & 92.4 & 0.737 & 11.2 & 87.8 & 0.796 & 21.4 & 73.9 & 1.263 & 29.1 & 57.6 & 2.123 & 22.5 & 69.5 & 1.574 & 3.43 & 1.39 & 1.50 \\
Cosmos-3 & \underline{6.8} & \underline{93.8} & \underline{0.676} & \underline{9.3} & \underline{90.5} & \underline{0.735} & \underline{15.1} & \textbf{83.5} & \textbf{0.957} & \underline{21.2} & \underline{75.5} & \underline{1.281} & \underline{16.6} & \underline{81.4} & \underline{1.060} & \underline{0.34} & \underline{0.22} & \textbf{0.67} \\
\rowcolor{blockbg}\method & \textbf{6.6} & \textbf{94.3} & \textbf{0.539} & \textbf{8.9} & \textbf{90.6} & \textbf{0.654} & \textbf{13.3} & \underline{83.1} & \textbf{0.957} & \textbf{17.4} & \textbf{76.5} & \textbf{1.259} & \textbf{14.2} & \textbf{81.9} & \textbf{1.027} & \textbf{0.23} & \textbf{0.10} & \textbf{0.67} \\
\specialrule{0.9pt}{1.5pt}{2pt}
\rowcolor{dsbg}\multicolumn{19}{@{}l@{}}{\textbf{\textsc{Cityscapes}}\hspace{6pt}\textcolor{blocktxt}{\scriptsize next 0.18\,s, short 0.35--0.53\,s, mid 0.71--0.88\,s}}\\[0.5pt]
Copy-last & 19.4 & 71.7 & 1.158 & 22.2 & 66.7 & 1.296 & 24.8 & 62.6 & 1.481 & -- & -- & -- & 22.7 & 66.1 & 1.342 & 1.83 & 1.22 & 0.34 \\
DINO-Foresight & 20.1 & 68.9 & -- & 21.1 & 67.1 & -- & 22.6 & 64.2 & -- & -- & -- & -- & 21.5 & 66.3 & -- & -- & -- & -- \\
VGGT-World & 19.3 & 70.8 & 1.228 & 22.1 & 67.5 & 1.322 & 25.7 & 62.8 & 1.577 & -- & -- & -- & 23.0 & 66.3 & 1.405 & 0.91 & 1.27 & 0.77 \\
Cosmos-3 & \underline{17.7} & \underline{74.6} & \underline{1.119} & \underline{18.7} & \underline{72.6} & \underline{1.158} & \underline{20.1} & \underline{70.6} & \underline{1.258} & -- & -- & -- & \underline{19.1} & \underline{72.2} & \underline{1.190} & \underline{0.08} & \underline{0.10} & \underline{0.31} \\
\rowcolor{blockbg}\method & \textbf{12.6} & \textbf{85.1} & \textbf{0.948} & \textbf{15.1} & \textbf{80.9} & \textbf{1.068} & \textbf{17.9} & \textbf{75.7} & \textbf{1.235} & -- & -- & -- & \textbf{15.7} & \textbf{79.7} & \textbf{1.111} & \textbf{0.02} & \textbf{0.03} & \textbf{0.14} \\
\specialrule{0.9pt}{1.5pt}{2pt}
\rowcolor{dsbg}\multicolumn{19}{@{}l@{}}{\textbf{\textsc{TartanAir}}\hspace{6pt}\textcolor{blocktxt}{\scriptsize next 1 frame, short 2--5, mid 6--10, long 11--20 frames}}\\[0.5pt]
Copy-last & 19.2 & 86.2 & 1.018 & 34.9 & 72.4 & 1.815 & 52.1 & 56.5 & 2.802 & 67.6 & 44.1 & 3.688 & 54.8 & 55.0 & 2.958 & 0.95 & 0.17 & 1.25 \\
VGGT-World & 19.9 & 84.6 & 1.137 & 38.2 & 68.4 & 2.088 & 54.0 & 46.3 & 3.980 & \underline{55.4} & 36.6 & 4.683 & 49.8 & 47.8 & 3.811 & 0.80 & 0.34 & 1.99 \\
Cosmos-3 & \underline{13.6} & \underline{88.9} & \underline{0.843} & \underline{20.9} & \underline{83.2} & \underline{1.175} & \underline{36.5} & \underline{69.6} & \underline{2.015} & 57.5 & \underline{51.3} & \underline{3.268} & \underline{42.8} & \underline{64.1} & \underline{2.415} & \underline{0.28} & \underline{0.14} & \underline{1.19} \\
\rowcolor{blockbg}\method & \textbf{12.0} & \textbf{89.8} & \textbf{0.786} & \textbf{18.6} & \textbf{84.5} & \textbf{1.094} & \textbf{28.9} & \textbf{73.3} & \textbf{1.844} & \textbf{40.1} & \textbf{57.3} & \textbf{3.151} & \textbf{31.6} & \textbf{68.3} & \textbf{2.294} & \textbf{0.13} & \textbf{0.09} & \textbf{0.77} \\
\specialrule{0.9pt}{1.5pt}{2pt}
\rowcolor{dsbg}\multicolumn{19}{@{}l@{}}{\textbf{\textsc{DOMINO, in-domain}}\hspace{6pt}\textcolor{blocktxt}{\scriptsize next 0.24\,s, short 0.48--0.96\,s, mid 1.92--2.88\,s, long 4.80--7.68\,s}}\\[0.5pt]
Copy-last & \underline{14.1} & \underline{85.3} & \underline{0.032} & \underline{17.1} & \underline{82.6} & \underline{0.040} & 21.6 & 77.7 & \underline{0.055} & 27.5 & \underline{73.6} & \underline{0.079} & 20.9 & 79.0 & \underline{0.054} & -- & -- & -- \\
VGGT-World & 16.5 & 83.7 & 0.041 & 20.3 & 79.7 & 0.050 & 27.1 & 73.5 & 0.072 & 30.6 & 67.1 & 0.090 & 24.6 & 74.9 & 0.066 & -- & -- & -- \\
Cosmos-3 & 15.2 & 85.1 & 0.040 & 17.6 & \underline{82.6} & 0.046 & \underline{20.8} & \underline{78.4} & 0.060 & \underline{25.8} & 73.3 & 0.081 & \underline{20.5} & \underline{79.1} & 0.059 & -- & -- & -- \\
\rowcolor{blockbg}\method & \textbf{4.4} & \textbf{95.0} & \textbf{0.016} & \textbf{7.2} & \textbf{91.1} & \textbf{0.025} & \textbf{10.8} & \textbf{84.7} & \textbf{0.040} & \textbf{13.4} & \textbf{81.1} & \textbf{0.051} & \textbf{9.6} & \textbf{87.0} & \textbf{0.036} & -- & -- & -- \\
\bottomrule
\end{tabular}
\end{table*}

\subsection{Implementation Details}
\label{sec:exp:impl}
While any feed-forward reconstruction model can serve as the backbone, we use a frozen VGGT-1B~\cite{vggt}, which returns depth, intrinsics, and poses for the observed frames in one pass.
Its depth is defined up to a scale per window, so during training the ground truth is brought to that scale with a factor computed from the last observed frame; on DOMINO, a single constant fitted on the training set converts the backbone's depth to metres, so the network predicts depth at its true scale.
Pixels without ground truth are masked, and ground-truth depth is capped at 80\,m on the driving datasets and 100\,m on TartanAir.
The pure-noise fraction of training samples is 0.4 and the depth floor $\dmin$ of the inverse-depth weight is 0.05; on TartanAir, where scenes are mostly near-range, we omit this weight so as not to suppress the sparse far structure.
We train with AdamW at a constant learning rate of $10^{-4}$, weight decay 0.01, batch size 16, gradient clipping at 1.0, and bf16 precision for 100k steps, and report results with an exponential moving average of the weights (decay 0.999).

\begin{table*}[t]
\caption{\textbf{Generalisation on DOMINO to unseen tasks and robots.} Same metrics, horizon groups, and ranking as the DOMINO block of \cref{tab:main}, which reports the seen tasks and robots.}
\label{tab:domino}
\centering
\fontsize{8.5}{10}\selectfont
\setlength{\tabcolsep}{2.4pt}
\renewcommand{\arraystretch}{1.04}
\newcolumntype{C}{>{\hspace{1.2pt}}c<{\hspace{1.2pt}}}
\begin{tabular}{@{}l ccC ccC ccC ccC ccC @{}}
\toprule
\multirow{2}{*}{Method} & \multicolumn{3}{c}{Next} & \multicolumn{3}{c}{Short} & \multicolumn{3}{c}{Mid} & \multicolumn{3}{c}{Long} & \multicolumn{3}{c}{All}  \\
\cmidrule(lr){2-4}\cmidrule(lr){5-7}\cmidrule(lr){8-10}\cmidrule(lr){11-13}\cmidrule(lr){14-16}
& AbsRel\,$\downarrow$ & $\delta_1$\,$\uparrow$ & CD\,$\downarrow$ & AbsRel\,$\downarrow$ & $\delta_1$\,$\uparrow$ & CD\,$\downarrow$ & AbsRel\,$\downarrow$ & $\delta_1$\,$\uparrow$ & CD\,$\downarrow$ & AbsRel\,$\downarrow$ & $\delta_1$\,$\uparrow$ & CD\,$\downarrow$ & AbsRel\,$\downarrow$ & $\delta_1$\,$\uparrow$ & CD\,$\downarrow$   \\
\midrule
\rowcolor{dsbg}\multicolumn{16}{@{}l@{}}{\textbf{\textsc{DOMINO, novel task}}}\\[0.5pt]
Copy-last & \underline{14.1} & 85.4 & \underline{0.032} & \underline{16.0} & 83.1 & \underline{0.039} & 22.6 & 77.6 & \underline{0.055} & 30.9 & 71.6 & \underline{0.075} & 21.9 & 78.6 & \underline{0.053} \\
VGGT-World & 16.0 & 84.1 & 0.040 & 18.7 & 81.1 & 0.050 & 27.4 & 73.6 & 0.071 & 32.3 & 65.9 & 0.079 & 24.7 & 75.0 & 0.063 \\
Cosmos-3 & 15.1 & \underline{85.9} & 0.038 & 16.6 & \underline{84.1} & 0.044 & \underline{22.3} & \underline{78.8} & 0.059 & \underline{27.9} & \textbf{73.3} & \underline{0.075} & \underline{21.2} & \underline{79.7} & 0.057 \\
\rowcolor{blockbg}\method & \textbf{5.1} & \textbf{93.7} & \textbf{0.017} & \textbf{7.4} & \textbf{90.3} & \textbf{0.026} & \textbf{14.7} & \textbf{81.1} & \textbf{0.048} & \textbf{23.4} & \underline{71.8} & \textbf{0.071} & \textbf{13.7} & \textbf{82.9} & \textbf{0.044} \\
\specialrule{0.9pt}{1.5pt}{2pt}
\rowcolor{dsbg}\multicolumn{16}{@{}l@{}}{\textbf{\textsc{DOMINO, novel embodiment}}}\\[0.5pt]
Copy-last & \underline{11.3} & 87.6 & \underline{0.030} & 13.4 & 84.4 & \underline{0.036} & 16.1 & 80.4 & \underline{0.047} & 20.9 & 75.5 & \underline{0.072} & 16.0 & 81.2 & \underline{0.049} \\
VGGT-World & 13.5 & 87.1 & 0.040 & 15.5 & 84.1 & 0.045 & 21.0 & 78.1 & 0.066 & 23.0 & 71.5 & 0.082 & 18.9 & 79.2 & 0.061 \\
Cosmos-3 & 11.6 & \underline{88.9} & 0.037 & \underline{13.0} & \underline{86.7} & 0.042 & \underline{15.1} & \underline{83.0} & 0.053 & \underline{19.7} & \underline{76.9} & 0.075 & \underline{15.3} & \underline{83.2} & 0.054 \\
\rowcolor{blockbg}\method & \textbf{4.4} & \textbf{94.6} & \textbf{0.018} & \textbf{6.5} & \textbf{91.4} & \textbf{0.026} & \textbf{9.5} & \textbf{85.6} & \textbf{0.041} & \textbf{12.7} & \textbf{80.3} & \textbf{0.056} & \textbf{8.8} & \textbf{87.0} & \textbf{0.038} \\
\specialrule{0.9pt}{1.5pt}{2pt}
\rowcolor{dsbg}\multicolumn{16}{@{}l@{}}{\textbf{\textsc{DOMINO, novel task and novel embodiment}}}\\[0.5pt]
Copy-last & \underline{10.9} & 88.4 & \underline{0.029} & 12.8 & 85.4 & \underline{0.036} & 15.7 & 80.4 & \underline{0.050} & 21.1 & \underline{72.4} & \underline{0.082} & 15.7 & 80.7 & \underline{0.052} \\
VGGT-World & 12.7 & 86.5 & 0.040 & 14.8 & 83.4 & 0.046 & 18.4 & 77.1 & 0.062 & 25.2 & 67.2 & 0.085 & 18.5 & 77.4 & 0.061 \\
Cosmos-3 & 11.1 & \underline{89.1} & 0.035 & \underline{12.4} & \underline{87.0} & 0.040 & \underline{14.8} & \underline{82.4} & 0.053 & \underline{19.6} & \textbf{74.8} & \underline{0.082} & \underline{14.9} & \underline{82.5} & 0.055 \\
\rowcolor{blockbg}\method & \textbf{4.8} & \textbf{93.6} & \textbf{0.018} & \textbf{6.5} & \textbf{90.8} & \textbf{0.026} & \textbf{10.8} & \textbf{83.1} & \textbf{0.048} & \textbf{18.3} & 72.0 & \textbf{0.080} & \textbf{10.9} & \textbf{83.6} & \textbf{0.046} \\
\bottomrule
\end{tabular}
\end{table*}

\subsection{Main Results}
\label{sec:exp:main}
\cref{tab:main} compares \method with a recent, strong representative of each paradigm in \cref{sec:related}: Cosmos-3~\cite{cosmos3}, a state-of-the-art video world model trained on large-scale video data, a 16B model that we run zero-shot as released and read out with the same frozen backbone, and VGGT-World~\cite{vggtworld}, a state-of-the-art latent world model, which forecasts in the latent space of that backbone and reaches long horizons by autoregressive rollout.
For VGGT-World we use the released checkpoints on KITTI and Cityscapes, and train it ourselves with the authors' model and recipe on TartanAir and DOMINO, for which no weights were released.
DINO-Foresight~\cite{dinoforesight}, which forecasts self-supervised image features, is listed for reference on the two driving datasets, and copy-last marks what no forecasting at all achieves.
\method achieves the lowest AbsRel at every horizon group and the lowest all-horizon CD on all four datasets among the ranked methods.
On all three moving-camera datasets, it also achieves the lowest ATE and RTE.

\noindent\textbf{Depth forecasting.}
\method has the lowest AbsRel at every horizon group on every dataset and the highest $\delta_1$ in all but one cell, and its margin grows with the horizon.
On KITTI it matches Cosmos-3 at the first frame (6.6 vs.\ 6.8, within the seed-to-seed variation of this split) and leads by four points at the long horizons (17.4 vs.\ 21.2); on TartanAir the lead grows from 1.6 points at the first frame to 17 points at the long horizons (40.1 vs.\ 57.5).
The rollout baseline shows the opposite trend: VGGT-World falls below copy-last from the mid horizons on KITTI and up to the mid horizons on TartanAir, which is the error accumulation that direct prediction avoids.
On DOMINO, \method cuts the error of every other method by about half at every horizon (9.6 vs.\ 20.5 for Cosmos-3 over all horizons).

\noindent\textbf{3D geometry forecasting.}
\begin{table}[t]
\caption{\textbf{Ablations on KITTI.}
Variants of \method, evaluated on the windows and with the protocol of \cref{tab:main}: median-scaled AbsRel~($\times 100$) and $\delta_1$~(\%), averaged over the KITTI horizon groups of that table, where \emph{All} averages every horizon.
$N$ is the number of noise draws averaged at read-out.}
\label{tab:ablation}
\centering
\fontsize{7.9}{9.3}\selectfont
\setlength{\tabcolsep}{3.2pt}
\renewcommand{\arraystretch}{0.95}
\resizebox{0.98\columnwidth}{!}{%
\begin{tabular}{@{}l cc cc cc@{}}
\toprule
\multirow{2}{*}{Variant} & \multicolumn{2}{c}{Short} & \multicolumn{2}{c}{Long} & \multicolumn{2}{c}{All} \\
\cmidrule(lr){2-3}\cmidrule(lr){4-5}\cmidrule(lr){6-7}
& AbsRel$\downarrow$ & $\delta_1$$\uparrow$ & AbsRel$\downarrow$ & $\delta_1$$\uparrow$ & AbsRel$\downarrow$ & $\delta_1$$\uparrow$ \\
\midrule
Copy-last & 13.9 & 82.6 & 22.5 & 71.8 & 19.2 & 75.8 \\
\specialrule{0.8pt}{1.1pt}{1.3pt}
\rowcolor{dsbg}\multicolumn{7}{@{}l@{}}{\textbf{\textsc{Prediction space}}}\\[-0.3pt]
feature space & 11.7 & 86.6 & 21.4 & 72.8 & 17.6 & 78.1 \\
feature space, + depth loss & 11.8 & 86.6 & 18.8 & 73.7 & 16.0 & 78.8 \\
\specialrule{0.8pt}{1.1pt}{1.3pt}
\rowcolor{dsbg}\multicolumn{7}{@{}l@{}}{\textbf{\textsc{Geometric conditioning}}}\\[-0.3pt]
no depth conditioning & 29.1 & 43.9 & 28.0 & 47.3 & 28.6 & 45.5 \\
uncompensated anchor & 10.6 & 87.6 & 17.8 & 75.9 & 15.1 & 80.3 \\
anchor, no history & 9.4 & 89.9 & 18.7 & 74.5 & 15.1 & 80.5 \\
last depth map only & 12.8 & 84.9 & 18.3 & 75.3 & 16.2 & 78.8 \\
\specialrule{0.8pt}{1.1pt}{1.3pt}
\rowcolor{dsbg}\multicolumn{7}{@{}l@{}}{\textbf{\textsc{Forecasting strategy}}}\\[-0.3pt]
one-step recursive rollout & 11.3 & 87.7 & 38.9 & 53.0 & 27.7 & 66.5 \\
two-step block rollout & 9.7 & 90.0 & 31.4 & 60.6 & 22.4 & 72.5 \\
\specialrule{0.8pt}{1.1pt}{1.3pt}
\rowcolor{dsbg}\multicolumn{7}{@{}l@{}}{\textbf{\textsc{Observed frames}}}\\[-0.3pt]
$\K = 2$ & 10.1 & 89.2 & 18.7 & 75.1 & 15.4 & 80.4 \\
$\K = 3$ & 9.4 & 89.9 & 17.7 & 76.0 & 14.5 & 81.3 \\
$\K = 4$ & 9.3 & 90.2 & 17.6 & 76.3 & 14.4 & 81.6 \\
\specialrule{0.8pt}{1.1pt}{1.3pt}
\rowcolor{dsbg}\multicolumn{7}{@{}l@{}}{\textbf{\textsc{Read-out}}}\\[-0.3pt]
Euler, 2 steps & 9.8 & 90.1 & 18.1 & 76.2 & 14.9 & 81.4 \\
Euler, 4 steps & 11.6 & 88.0 & 19.7 & 74.0 & 16.6 & 79.2 \\
Euler, 8 steps & 14.0 & 84.2 & 22.4 & 70.1 & 19.2 & 75.4 \\
\specialrule{0.8pt}{1.1pt}{1.3pt}
\shortstack[l]{\method\\[-1.5pt]{\scriptsize($\K{=}5$, anchor, $N{=}1$)}} & 9.0 & 90.5 & 17.5 & \textbf{76.5} & \textbf{14.2} & 81.8 \\
\rowcolor{blockbg}\shortstack[l]{\method\\[-1.5pt]{\scriptsize($\K{=}5$, anchor, $N{=}4$)}} & \textbf{8.9} & \textbf{90.6} & \textbf{17.4} & \textbf{76.5} & \textbf{14.2} & \textbf{81.9} \\
\bottomrule
\end{tabular}%
}
\end{table}

The Chamfer distance follows the depth results: \method has the lowest CD at every horizon group on all four datasets, with one tie against Cosmos-3 at the mid horizons of KITTI.
Over all horizons, the margin over Cosmos-3 ranges from 3\% on KITTI to 39\% on DOMINO.

\noindent\textbf{Camera-motion forecasting.}
On the three datasets with a moving camera, \method has the lowest trajectory error on all three metrics, with a single tie against Cosmos-3 on the rotation error of KITTI.
Its translation errors are clearly lower than those of Cosmos-3 (ATE 0.23 vs.\ 0.34 on KITTI, 0.02 vs.\ 0.08 on Cityscapes, 0.13 vs.\ 0.28 on TartanAir) and several times lower than those of VGGT-World, whose relative errors are worse than a static camera on all three datasets.
This accuracy comes from a predictor with a few tens of thousands of parameters.

\subsection{Generalisation to Novel Tasks and Robots}
\label{sec:exp:ood}
\cref{tab:domino} tests what happens when the task, the robot, or both are unseen during training.
\method has the lowest AbsRel and the lowest Chamfer distance in every cell, and it predicts the first horizon with an AbsRel between 4.4 and 5.1 where every other method is above 10.
The rows differ in difficulty, so we read the results relative to copy-last: over all horizons, \method reduces its error by 54\% on seen tasks and robots, by 45\% on an unseen robot, by 37\% on unseen tasks, and by 31\% when both are unseen.
An unseen robot therefore costs little, whereas unseen tasks cost more.
At the longest horizon of the two novel-task blocks Cosmos-3 has the higher $\delta_1$, and when both task and robot are unseen the AbsRel margin narrows to one point.

\subsection{Ablation Studies}
\label{sec:exp:ablation}
\begin{figure}[t]
  \centering
  \captionsetup{font=footnotesize,skip=4pt}
  \includegraphics[width=\columnwidth]{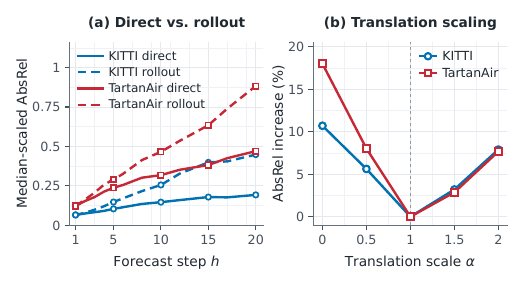}
  \caption{\textbf{Direct forecasting and geometric conditioning.}
  (a) Direct prediction versus one-step recursive rollout with the same checkpoint.
  (b) Increase in horizon-averaged AbsRel when scaling the anchor's predicted translation relative to $\alpha=1$; rotation is unchanged.
  Comparisons use matched window subsets and equal-drive averaging on KITTI.}
  \label{fig:probes}
\end{figure}

\begin{figure*}[t]
  \centering
  \captionsetup{font=footnotesize,skip=4pt}
  \includegraphics[width=\textwidth]{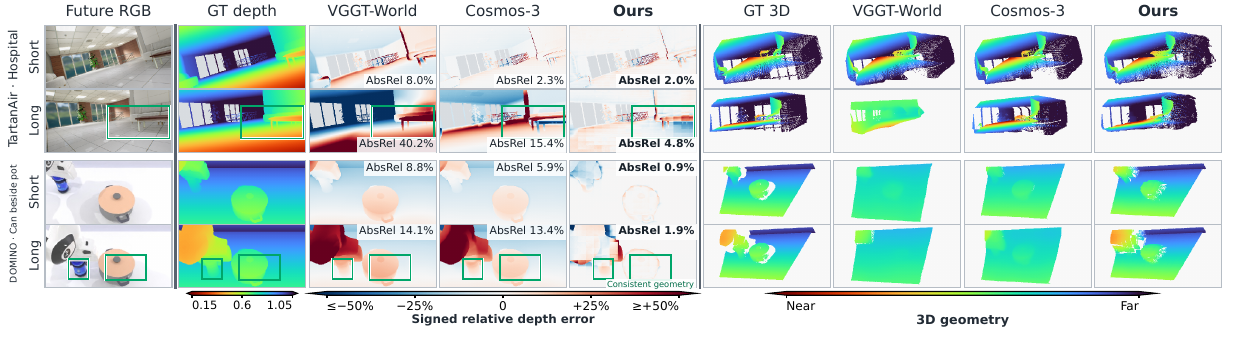}
  \caption{\textbf{Qualitative forecasts.}
  Short/long horizons are $h=1/10$ for TartanAir and $h=4/48$ for DOMINO.
  Error maps show signed relative depth error (white is zero); boxes mark corresponding scene regions.
  Clouds use calibrated intrinsics, median-scaled TartanAir depths and fixed-scale DOMINO depths; colors indicate depth.
  All clouds share the same view without further alignment.}
  \label{fig:qualitative}
\end{figure*}

\cref{tab:ablation} varies the design choices of \method on KITTI, with the evaluation windows and the protocol of \cref{tab:main} held fixed.

\noindent\textbf{Forecasting in geometry space.}
To test the first pillar of the method we train a twin of \method that keeps our architecture, our conditioning and our training schedule, but regresses the depth-head tokens of the frozen front end instead of depth, and decodes them back to depth with that same frozen head.
Its targets are the four depth-head layers of that front end, reduced by a per-layer PCA to 128 standardised components each.
The second variant is the stronger one: it adds a loss on the decoded depth, which aligns its objective with the evaluation metric and reduces its error over all horizons from 17.6 to 16.0, though not at the first frame.
Forecasting in explicit geometry is better than both variants at every horizon group, by at least 2.8 points at the short horizons, 1.4 at the long ones and 1.8 over all horizons, and it uses half as many tokens per frame.

\noindent\textbf{Geometric conditioning.}
Removing the depth conditioning altogether leaves an error worse than copy-last and almost unchanged by how far ahead the model is asked to predict.
The anchor and the observed depth history matter at different ranges.
Replacing the camera-compensated anchor by a plain copy of the last observed depth map costs most at the short horizons, whereas keeping the anchor and dropping the history costs most at the long ones.
Using only the last observed depth map removes both, and is worse than removing either one alone at the short horizons and over all horizons.
\cref{fig:probes}(b) asks whether the forecast uses the particular viewpoint the anchor encodes, or merely the presence of a warped depth map.
Rebuilding the anchor with the predicted translation scaled by a factor $\alpha$, rotation unchanged, traces a V whose minimum sits at $\alpha{=}1$, the predicted motion itself; removing the translation entirely costs 10.7\% on KITTI and 18.0\% on TartanAir.
A model that ignored the viewpoint would trace a flat curve instead.

\noindent\textbf{Direct prediction.}
The second pillar is tested by feeding the model its own predictions back into the context, with the same weights and the same windows.
The first step is identical by construction, and the two strategies separate immediately afterwards: at the long horizons a one-step recursive rollout reaches 38.9 AbsRel against 17.4 for direct prediction, and falls behind copy-last.
Predicting two steps from each context halves the number of feedback stages and recovers part of the gap without closing it.
\cref{fig:probes}(a) shows the same behaviour over the full horizon range on two datasets, with rollout ending at 2.3 times the direct error on KITTI and 1.9 times on TartanAir.

\noindent\textbf{Observed frames.}
Accuracy improves from two observed frames to four at every horizon group, and most of the gain is already reached at three.
We use $\K = 5$ in all other experiments.

\noindent\textbf{Read-out.}
One-step prediction at $\ft = 0$ outperforms Euler integration at every horizon group, which supports the choice made in \cref{sec:training}.
The degradation grows with the number of integration steps.
Averaging four noise draws rather than one gives a small further gain at the short and long horizons, and leaves the all-horizon error unchanged at the precision reported here.

\subsection{Qualitative Analysis}
\label{sec:exp:qual}
\cref{fig:qualitative} shows forecasts on TartanAir and DOMINO at short and long horizons, as signed relative depth-error maps, white at zero error, and as the resulting point clouds.
In the hospital scene, \method preserves the room layout more faithfully as the viewpoint changes, while baseline errors grow around the furniture and structural boundaries.
On DOMINO, it better retains the raised geometry of the can and pot, which the baselines tend to flatten, although object boundaries remain smooth.

\begin{table}[t]
\caption{\textbf{Inference cost.}
Compute, parameters and the time to obtain the forecasts of different horizons.}
\label{tab:cost}
\centering
\fontsize{7.0}{8.4}\selectfont
\setlength{\tabcolsep}{1.0pt}
\renewcommand{\arraystretch}{1.04}
\begin{tabular}{@{}l c ccc cccc@{}}
\toprule
\multirow{2}{*}{Method} & \multirow{2}{*}{TFLOPs} & \multicolumn{3}{c}{Params} & \multicolumn{4}{c}{Time to forecast (s)} \\
\cmidrule(lr){3-5}\cmidrule(lr){6-9}
& & forecaster & backbone & total & $\hz{=}1$ & $\hz{=}5$ & $\hz{=}10$ & $\hz{=}20$ \\
\midrule
DINO-Foresight~\cite{dinoforesight} & 30.6 & 256\,M & DINO & 361\,M & 0.03 & 0.11 & 0.22 & 0.42 \\
VGGT-World~\cite{vggtworld}         & 1116 & 433\,M & VGGT & 1.59\,B & 1.9 & 7.7 & 17.4 & 36.9 \\
Cosmos-3~\cite{cosmos3}             & 3362 & 17.0\,B & VGGT & 18.1\,B & 4.9 & 5.8 & 6.7 & 9.3 \\
\rowcolor{blockbg}\method                      & 12.0 & 206\,M & VGGT & 1.36\,B & 0.11 & 0.11 & 0.11 & 0.11 \\
\bottomrule
\end{tabular}
\end{table}

\subsection{Computational Cost}
\label{sec:exp:cost}
\cref{tab:cost} compares inference cost.
A horizon is an input to \method rather than a number of steps to take, so the same four-draw read-out answers any horizon and the latency is flat at 0.11\,s across the evaluated range; producing all twenty horizons costs 12\,TFLOPs.
VGGT-World and DINO-Foresight instead extend their forecasts through sequential rollout, so VGGT-World grows from 1.9\,s to 36.9\,s and spends 1116\,TFLOPs over the same twenty horizons.
Cosmos-3 does not roll out: it denoises a whole clip at once with a 17\,B generator, which requires 4.9\,s even at the nearest evaluated horizon.
\method therefore reaches the longest horizon over 300$\times$ faster than VGGT-World and over 80$\times$ faster than Cosmos-3, with the smallest forecaster of the four and the same frozen backbone as VGGT-World.

\section{Conclusion}
\label{sec:conclusion}
We presented \method, a world model that forecasts directly in explicit geometry.
\method outperforms the evaluated video- and feature-based world models, while representation ablations further support predicting geometry directly.
Direct horizon conditioning further avoids recursive feedback and its associated error accumulation, enabling efficient prediction at longer horizons.
A lightweight camera-motion predictor improves the forecast by relating observed geometry to the future viewpoint.
Across driving, aerial flight, and robotic manipulation, \method produces accurate future geometry and camera-motion forecasts, while generalising to unseen manipulation tasks and robot embodiments.
Together, these results establish \method as an accurate and efficient approach to future geometry forecasting and demonstrate the value of combining explicit geometric prediction with direct horizon conditioning.
Conditioning the forecast on robot actions is a natural follow-up.
Additionally, predicting motion and geometry jointly is another direction for future work.

\bibliographystyle{IEEEtran}
\bibliography{refs}

\end{document}